\documentclass[letterpaper, 10 pt, conference]{ieeeconf}  

\IEEEoverridecommandlockouts                              

\usepackage[english]{babel}
\usepackage[utf8]{inputenc}
\usepackage{svg}

\usepackage{amsfonts,amssymb,amsmath}
\newtheorem{problem}{Problem}

\usepackage{algorithm}
\usepackage{algorithmic}

\usepackage{graphicx}
\usepackage{subfigure}
\usepackage{subcaption}

\usepackage{todonotes}

\newcommand{\eg}{\textit{e.g.}}
\newcommand{\ie}{\textit{i.e.}}
\newcommand{\etal}{\textit{et al.}}

\title{\LARGE \bf
Beyond the Plane: A 3D Representation of Human Personal Space\\for Socially-Aware Robotics
}

\author{Caio C. G. Ribeiro \qquad Douglas G. Macharet
\thanks{This work was supported by CAPES/Brazil - Finance Code 001, CNPq/Brazil, and FAPEMIG/Brazil.}
\thanks{The authors are with the Computer Vision and Robotics Laboratory (VeRLab), Department of Computer Science, Universidade Federal de Minas Gerais, Brazil. E-mails: {\tt\small caioconti@ufmg.br, doug@dcc.ufmg.br}.}
}
\begin{document}
\onecolumn
    {\Large \textbf{IEEE Copyright Notice}}\\[2em]
    
    © 2025 IEEE. Personal use of this material is permitted. Permission from IEEE must be obtained for all other uses, in any current or future media, including reprinting/republishing this material for advertising or promotional purposes, creating new collective works, for resale or redistribution to servers or lists, or reuse of any copyrighted component of this work in other works.\\[1.5em]
    
    \textit{Accepted to be published in: 2025 34th IEEE International Conference on Robot and Human Interactive Communication (RO-MAN).}

\twocolumn
\maketitle
\thispagestyle{empty}
\pagestyle{empty}


\begin{abstract}

The increasing presence of robots in human environments requires them to exhibit socially appropriate behavior, adhering to social norms. A critical aspect in this context is the concept of personal space, a \emph{psychological boundary} around an individual that influences their comfort based on proximity. This concept extends to human-robot interaction, where robots must respect personal space to avoid causing discomfort. 
While much research has focused on modeling personal space in two dimensions, almost none have considered the vertical dimension. 
In this work, we propose a novel three-dimensional personal space model that integrates both height (introducing a discomfort function along the Z-axis) and horizontal proximity (via a classic XY-plane formulation) to quantify discomfort. To the best of our knowledge, this is the first work to compute discomfort in 3D space at any robot component’s position, considering the person’s configuration and height.
\end{abstract}


\section{Introduction}

The concept of personal space plays a crucial role in social robotics, defining the boundaries within which a human feels comfortable with another person or robot nearby. As robots become increasingly integrated into human environments~\cite{Mavrogiannis2023}
understanding personal space -- also referred to as \emph{proxemics}~\cite{Hall1966} -- becomes essential. Respecting these spatial boundaries enables socially aware and acceptable robot navigation, fostering a more comfortable and natural coexistence between humans and robots.


In this context, several researchers have proposed methods to compute the personal space~\cite{Kirby2010, Hoang2023, Manuela2024}, aiming to quantify human comfort levels. While no universally accepted approach exists~\cite{Mavrogiannis2023}, most methods define personal space in two dimensions, typically on the ground plane. This simplification aligns well with socially aware navigation, as many mobile robots can be approximated as points or simple shapes moving in a planar environment.



\begin{figure}[t]
    \centering
        \centering
        \includegraphics[width=0.85\linewidth]{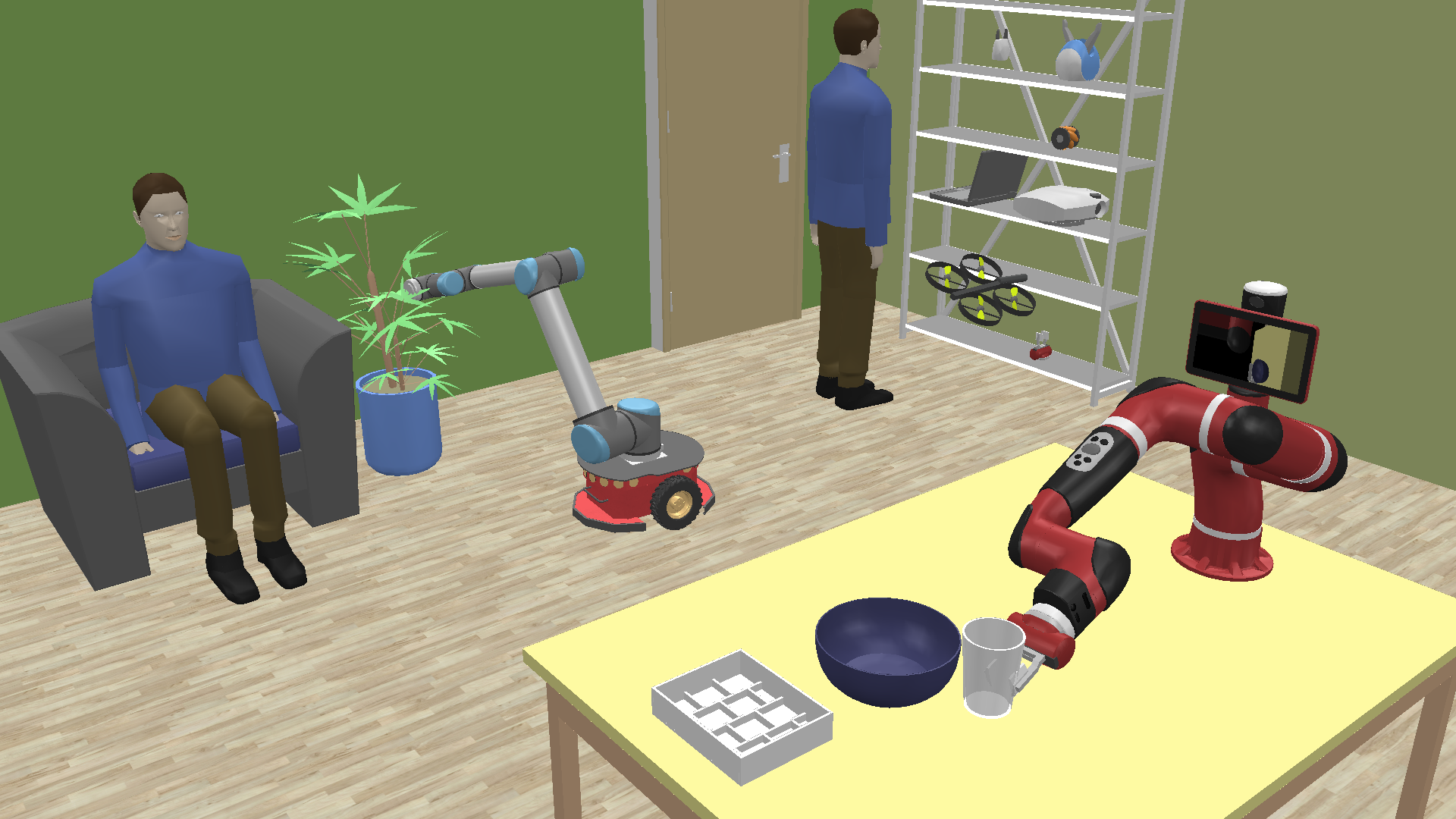}  
    \caption{
    Example: A mobile manipulator performing gardening tasks, a drone inspecting a shelf, and a manipulator handling an object -- each of these operations occurs in 3D around humans and may impact their comfort level.
    }
    \label{fig:initial_figure}
\end{figure}

However, existing techniques have limitations when applied to robots that operate in the three-dimensional space, such as drones and mobile manipulators (Figure.~\ref{fig:initial_figure}). Traditional 2D approaches restrict comfort computation to the XY plane, neglecting height (Z-axis) and potentially leading to undesirable behaviors, \eg, moving near the head of a person. As the presence of drones and mobile manipulators in human environments continues to grow~\cite{Garrell2017, Thakar2022}, the need for a comprehensive three-dimensional personal space model becomes increasingly important.

Despite this need, research on three-dimensional personal space remains limited. Most existing studies focus on drones~\cite{Garrell2017, Bretin2024}, but since drones primarily operate above humans, their findings do not generalize to a complete 3D proxemic framework. In contrast, ground-based mobile manipulators and fixed manipulators interact with humans at different heights, posing distinct challenges in defining personal space. Notably, the concept of 3D proxemics extends to flying robots operating below head level, assuming that humans exhibit similar comfort responses to different types of robots in comparable spatial conditions. This motivates the development of a more generalizable three-dimensional personal space representation.


In this work, we propose a novel three-dimensional personal space representation (3D Proxemics) given the increasing presence of robots in 3D environments and the lack of dedicated research on this topic. Our approach incorporates the concept of Maximum Permissible Pressure (MPP) from established robotic norms~\cite{ISO} and employs Fuzzy Logic to model a smooth comfort gradient along the Z-axis. By integrating this with a traditional formulation for personal space in the XY plane, we achieve a comprehensive 3D model. The resulting model quantifies a person’s comfort level at any point in 3D space, ranging from 0 (completely comfortable) to 1 (completely uncomfortable).

In summary, our contributions are as follows:
\begin{itemize} 
    \item Proposition of a comfort level along the Z-axis (height) based on the Maximum Permissible Pressure concept; 
    \item A method for integrating the proposed Z-axis comfort level with a XY-plane personal space formulation; 
    \item A novel three-dimensional personal space representation for the full range of possible human-robot interactions. 
\end{itemize}

\section{Related Work}


In social robotics, robots must navigate safely and acceptably around humans by considering not only obstacle avoidance but also social norms~\cite{Mavrogiannis2023}.
To achieve this, previous research has focused on developing socially-aware robots by incorporating social cues into planning frameworks \cite{Silva2023, Aline2023}. A fundamental social norm for robots to respect is human personal space, and its accurate representation is key for positive human-robot interactions.



Since their introduction in psychology by Hall~\cite{Hall1966}, numerous attempts have been made to define personal spaces, or proxemics, which describe the distances at which humans feel most comfortable interacting with others, including robots~\cite{Narayanan2023}. Initially modeled as concentric circular zones~\cite{Hall1966}, personal spaces were later refined using Gaussian functions to form egg-shaped regions that more accurately reflect human spatial boundaries~\cite{Kirby2010}. This shape continues to be used -- with variations -- in recent works~\cite{Narayanan2023, Aline2023, Hoang2023}.

However, these approaches model personal space only in two dimensions (XY plane), neglecting the vertical component. While suitable for ground-based robots operating on planar surfaces, a three-dimensional representation is essential for aerial robots, robotic arms, and mobile manipulators to ensure socially-aware interaction in the vertical dimension.


%

Several studies have explored human-drone proxemics with varying results. Duncan and Murphy~\cite{Duncan2013} found no significant preference regarding drone height, a finding echoed by Kunde~\etal~\cite{Kunde2022}.
However, Yeh~\etal~\cite{Yeh2017} observed people approaching drones more closely at 1.2 meters compared to 1.8 meters, while Bevins~\etal~\cite{Bevins2024} found that the preferred height ranged from chest to eye level. In contrast, Ariyasena~\etal~\cite{Ariyasena2020} and Bretin~\etal~\cite{Bretin2024} reported a preference for drones at higher altitudes, typically 1.9 to 3 meters or above eye level, depending on the task. Despite these findings, none of these studies propose a mathematical formulation to quantify discomfort along the vertical axis or in a full three-dimensional personal space, as is common in existing 2D models.


Garrell~\etal~\cite{Garrell2017} is the only study found that proposes a three-dimensional personal space model for drone accompaniment tasks. 
However, it assumes the 3D model as a simple extension of the 2D version, which contrasts with our approach that defines a novel Z-axis formulation and integrates it with an existing 2D model.
Beyond aerial robots, few studies have addressed 3D comfort modeling. The closest approach is Sisbot~\etal~\cite{Sisbot2012}, which introduces a Safety Criterion in 3D space for socially-aware mobile manipulation. Their model focuses on safety by increasing distance from vulnerable body regions, while our model incorporates sensitivity data from the entire body and emphasizes co-existence rather than direct interaction.

We propose a novel method for computing three-dimensional personal space, with a focus on robot navigation that prioritizes coexistence by minimizing discomfort to humans, rather than being tailored to specific interactions.
To the best of our knowledge, no prior studies have quantified comfort in three dimensions, either in psychology or robotics. While some works recognize that certain areas, such as the head and face, are more sensitive and evoke greater stress in individuals, they have not fully explored the three-dimensional nature of personal space.

\section{Problem Formulation}

Let $\mathcal{E} \in \mathbb{R}^3$ represent an obstacle-free, human-populated environment.
Consider a given individual $\mathcal{P}$ with height $h$, described by a configuration $\mathbf{q} = (\mathbf{p}, \theta)$, where $\mathbf{p} = (x, y)$ denotes the position, and $\theta$ represents the orientation.

An individual $\mathcal{P}$ has an associated personal space $\mathcal{S} \subset \mathcal{E}$. 
The personal space $\mathcal{S}$ is a function that computes the discomfort value, in the range $[0,1]$, that an individual with configuration $\mathbf{q}$ and height $h$ experiences at a given point $r \in \mathbb{R}^3$. In other words, it answers the question: \emph{How much discomfort would I feel with the robot at that location?} Therefore:
\begin{equation}
\mathcal{S}(\mathcal{P}, \mathbf{r}) \in [0,1] ~.
\end{equation}



The problem consists of defining a comprehensive three-dimensional personal space representation that quantifies the discomfort caused by proximity to the individual $\mathcal{P}$ at any location $\mathbf{r}$ in the environment

More formally, our problem can be defined as:
\begin{problem}[Three-dimensional Personal Space] 

Let $\mathcal{E} \subset \mathbb{R}^3$ be an obstacle-free environment, and let $\mathcal{P}$ represent an individual within this environment.
The objective is to compute the personal space $\mathcal{S}(\mathcal{P}, \mathbf{r}) \in [0,1]$, \ie, a discomfort value, for any given point $\mathbf{r} \in \mathcal{E}$. The discomfort value should reflect the individual’s comfort level relative to their proximity to the point $\mathbf{r}$, with $0$ indicating no discomfort (maximum comfort) and $1$ indicating maximum discomfort (minimum comfort).
\end{problem}

\section{Methodology}

This section outlines the derivation of the 3D Proxemics equation in three key steps. First, we define discrete discomfort values for specific body regions using a proposed segmentation, mapping each value to its corresponding height relative to the individual's total height. Second, we apply a Fuzzy Inference System (FIS) to transform these discrete values into a continuous Z-axis comfort function, capturing the relationship between height and discomfort. Finally, we integrate this function with a XY-plane personal space model, forming a complete 3D discomfort representation.

\subsection{Body Discomfort-related Segmentation}


To the best of our knowledge, no comprehensive studies have explicitly quantified human discomfort based on an object's or robot's position along the Z-axis (height). However, existing findings suggest that proximity to the face or head induces greater discomfort~\cite{Rubagotti2022, Garrell2017}, implying a preference for external elements to be at lower regions.


Given the limited research on this topic, our approach builds on the established link between safety and comfort~\cite{Vischer2007}, where safety is a prerequisite for comfort. Prior studies in robotics identify varying levels of vulnerability across body regions, with discomfort or pain arising when pressure exceeds a certain threshold. Therefore, to promote safe and comfortable interactions, robots should maintain greater distances from these more sensitive areas.


The international standard ISO/TS 15066:2016~\cite{ISO}, revised in 2022, establishes safety guidelines for collaborative robots, defining precise pressure and force limits across different body regions to prevent minor injuries. In this work, we leverage the Maximum Permissible Pressure (MPP) as a proxy for sensitivity -- lower MPP values indicate more sensitive areas, implying that robots should maintain greater distances from these regions to enhance human comfort.



\begin{figure}[htpb]
    \centering
    \includegraphics[width=.9\linewidth]{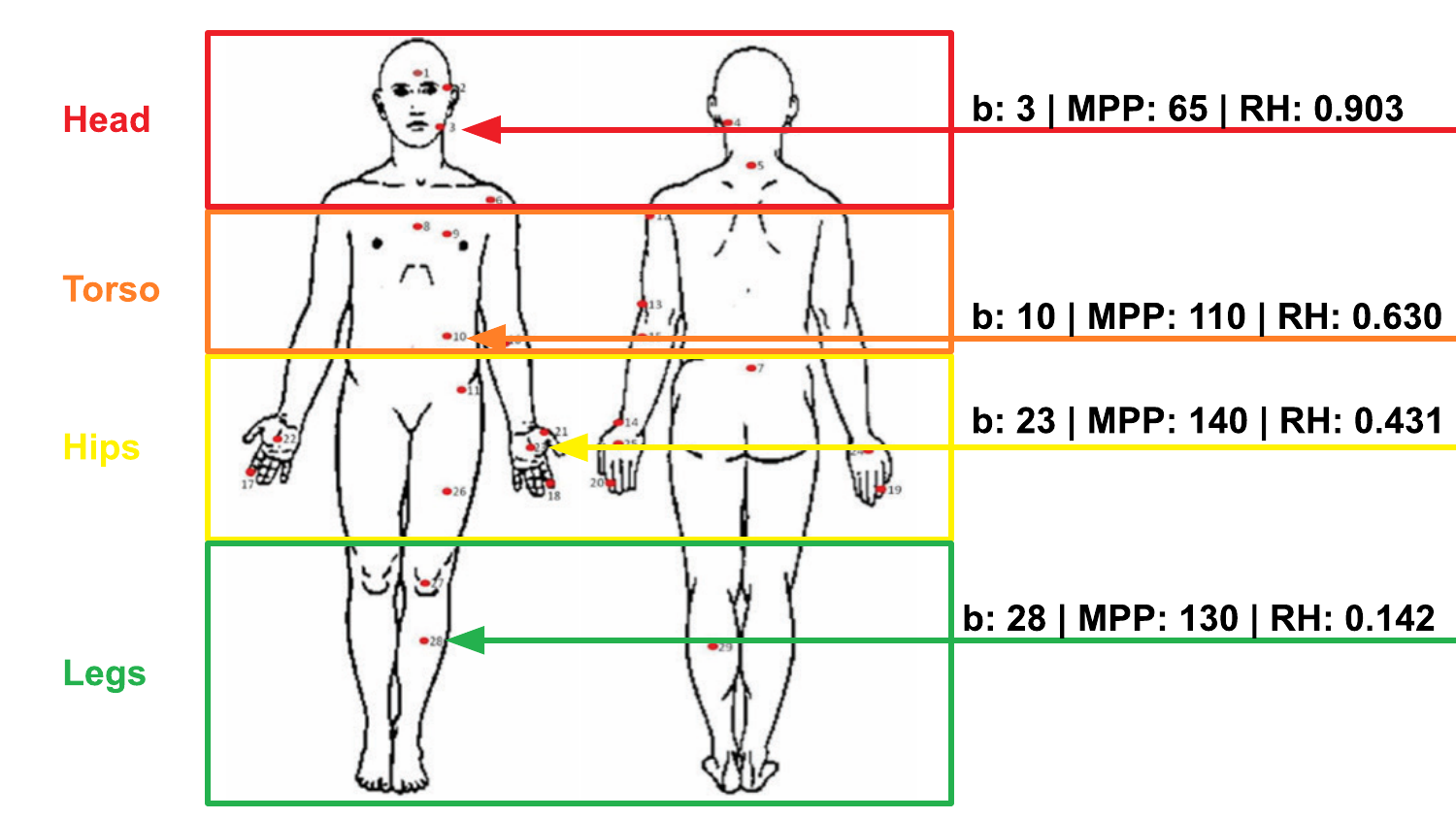}
    \caption{Adapted diagram from \cite{ISO}, with our segmentation, highlighting the most sensitive area $b$ within each body region, along with its corresponding Maximum Permissible Pressure~(MPP) and relative height~(RH).}
    \label{fig:body_regions}
\end{figure}

Since MPP values can vary significantly across closely spaced points, we first segment the human body into broader regions, selecting the lowest (most sensitive) MPP within each region. Following Haddadin~\etal~\cite{Haddadin2013}, who categorize safety-measured regions into parts such as the head, neck, torso, and lower extremities, we adopt a similar approach, dividing the body into regions based on spatial proximity.
We define four body regions: head, torso, hips, and legs, as illustrated in Figure~\ref{fig:body_regions}. The figure also highlights the most sensitive area $b$ within each region, along with its corresponding MPP value in N/cm².

For generalization, it is necessary to determine the location of each selected point relative to an individual's height. Using body segment proportions from Winter~\etal~\cite{Winter2009}, expressed as fractions of the height, we estimate the relative height ($RH$) of each chosen point. These proportions offer a reliable approximation when only the individual's height is available~\cite{Winter2009}. Figure~\ref{fig:body_regions} shows the results, where each $RH$ value represents a fraction of the height. For example, a body area with $RH = 0.903$ for a person with a height of $1.75\,m$ corresponds to a point located at $0.903 * 1.75 \approx 1.58\,m$.


Finally, we define the relationship between MPP and the degree of discomfort. As noted earlier, discomfort is inversely proportional to MPP, meaning that lower MPP values correspond to higher discomfort levels. To map the MPP values of each selected body area $b$ to discomfort, we divide the lowest MPP value (65, for the head) by the MPP of each region, ensuring the most sensitive point corresponds to a maximum discomfort value of 1.0 and the higher the value, the lower it gets.
%
%



We also define two additional regions outside the human height boundaries: the \emph{ground} and the \emph{top}. The ground is set at height 0.0, and the top is defined as $h + 0.75\,m$. For the ground, we assign a discomfort value of 1, as we aim to avoid robots making contact with the feet or the floor. For the top, we assign a discomfort value of 0, since there are no body parts above this height, and the robot's proximity in the Z-axis becomes negligible, thus no discomfort is caused.


The information for each selected body part $b$ and the two additional regions is summarized in Table~\ref{tab:parameters_relation}.
%
\begin{table}[htpb]
\centering
\caption{Parameters by regions, including Maximum Permissible Pressure (MPP), discomfort, and relative height (RH) as a proportion of the person’s height $h$.}
\begin{tabular}{|c|l|l|l|l|}
\hline
\multicolumn{1}{|l|}{\textbf{$b$}} & \textbf{Region} & \textbf{MPP} & \textbf{Discomfort} & \textbf{RH} \\ \hline
\textbf{3}                               & Head                 & 65           & 1.0                 & $0.903h$                    \\ \hline
\textbf{10}                              & Torso                & 110          & 0.591               & $0.630h$                    \\ \hline
\textbf{23}                              & Hips                 & 140          & 0.464               & $0.431h$                    \\ \hline
\textbf{28}                              & Legs                 & 130          & 0.500               & $0.142h$                    \\ \hline
\textbf{-}                              & Ground                 & -          & 1.0               & 0.0                    \\ \hline
\textbf{-}                              & Top                 & -          & 0.0               & $h + 0.75\,m$                    \\ \hline
\end{tabular}
\label{tab:parameters_relation}
\end{table}



\subsection{Fuzzy Inference System}


We use a Fuzzy Inference System (FIS) to derive a continuous equation that generalizes the relationship between discomfort and height, based on the six discrete data points described earlier. Fuzzy Inference Systems are effective at modeling human expertise (in this case, discomfort values) and are widely used as universal approximators~\cite{Jang1997, Zhao2023}. Our FIS ensures smooth transitions between the discrete regions, generating a continuous function that accurately captures comfort variations along the Z-axis.


For this task, we employ a Zero-order Sugeno FIS~\cite{Jang1997}, utilizing primarily Gaussian-type Membership Functions (MFs) for the antecedents and constant values for the consequents. The system has a single input variable $z$, \ie, the height, and produces a single output: discomfort ($[0, 1]$). We define six fuzzy sets to represent the regions: Legs, Hips, Torso, Head, Ground, and Top. Defuzzification is performed using a weighted average of all rule outputs, ensuring a smooth transition across the defined regions.



\begin{figure}[htpb]
    \centering
    \includegraphics[width=0.8\linewidth]{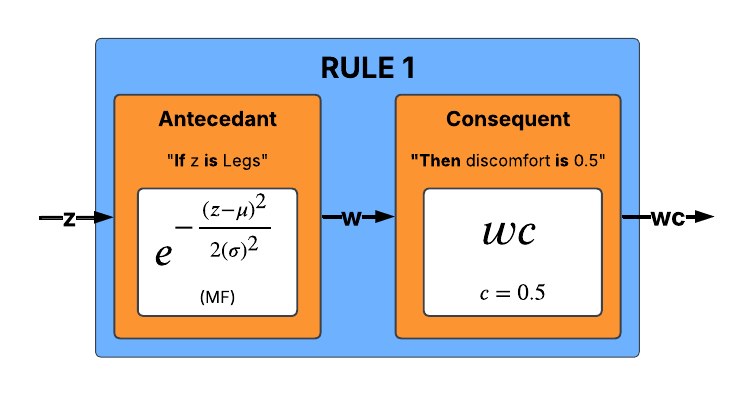}
    \caption{Diagram demonstrating how fuzzy rules operate.}
    \label{fig:rules_diagram}
    \vspace{-4mm}
\end{figure}

Rules are logical IF-THEN statements that define the relationship between the antecedent and the consequent. An example of a rule diagram is shown in Figure~\ref{fig:rules_diagram}. The antecedent corresponds to the IF part, and the consequent corresponds to the THEN part. We have formulated one rule for each fuzzy set, with the consequent of each rule representing the discomfort values, 
as the following:
%
%
\begin{enumerate}
    \item If $z$ is \textbf{legs}, then discomfort is $0.500$.
    \item If $z$ is \textbf{hips}, then discomfort is $0.464$.
    \item If $z$ is \textbf{torso}, then discomfort is $0.591$.
    \item If $z$ is \textbf{head}, then discomfort is $1.000$.
    \item If $z$ is \textbf{ground}, then discomfort is $1.000$.
    \item If $z$ is \textbf{top}, then discomfort is $0.000$.
\end{enumerate}

Given the rules, the first step is to determine the degree to which the input $z$ belongs to a specific region, known as the \emph{is region} part of the rule. In a FIS, the input has a membership degree for each fuzzy set, ranging from 0 (no membership) to 1 (full membership). In other words, the farther the input is from a region, the lower the degree of membership. This process is achieved using Membership Functions (MFs), which allow the system to handle uncertainty and imprecision, enabling smooth transitions between fuzzy sets and producing a more flexible and adaptive model for representing discomfort levels.

To model the degree of membership of an input $z$ to each of the six regions, we define six different MFs, one for each region. We employ Gaussian functions to model these MFs, given by the following equation:
\begin{equation}
    G_k(z) = e^{-\frac{(z - \mu_k)^2}{2(\sigma_k)^2}} ~,
    \label{eq:gaussian}
\end{equation}
where $\mu_k$ is the center of the Gaussian distribution and $\sigma_k$ is the standard deviation of the $k$-th fuzzy set. Therefore, $G_k(z)$ represents the degree of membership of input $z$ to the $k$-th region. The center $\mu_k$ of each Gaussian is set to the relative height corresponding to that region, as shown in Table~\ref{tab:parameters_relation}. The standard deviation $\sigma_k$ is set to $0.3$ for all body regions except head, $0.25$ for head, $0.1$ for ground and $0.3$ for top. These values were empirically fine-tuned to ensure smooth transitions between fuzzy sets.

\begin{figure}[htpb]
    \centering
    \includegraphics[width=0.95\linewidth]{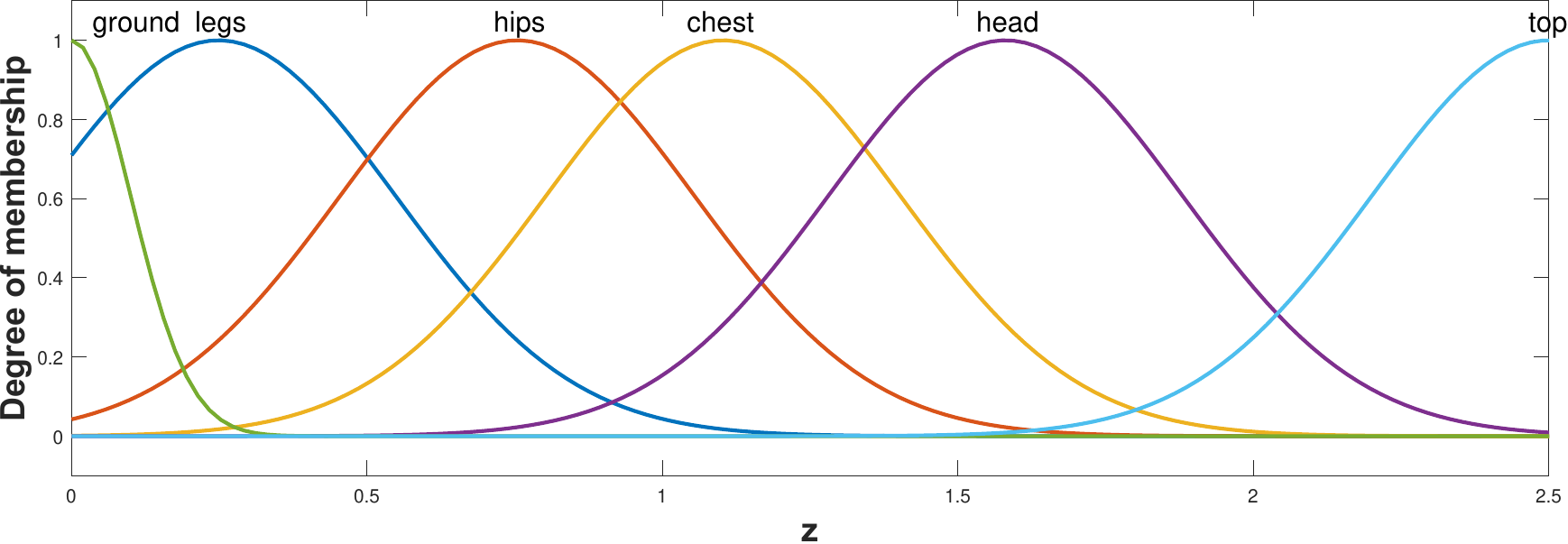}
    \caption{Visualization of the input Membership Functions (MFs) for an individual with a height of $h = 1.75\,m$.}
    \label{fig:MFs}
    \vspace{-2mm}
\end{figure}


With the MFs established and no use of AND/OR operators in our rules, the antecedent consists of a single condition: If $z$ is in a given region. Consequently, the output of the MF directly determines the antecedent ($w$), which is then used to weight the consequent of each rule without additional computations. This process is illustrated in Figure~\ref{fig:rules_diagram}.


Since our consequents are constant discomfort values $c$ (zero-order system), the rule output is computed as a simple product of the antecedent activation level and the consequent. These rule outputs are then aggregated and processed by the defuzzifier. The defuzzification step employs a weighted average, producing the final discomfort value as follows:
\begin{equation}
    f(z) = \frac{\sum_{i=1}^{n} w_i c_i}{\sum_{i=1}^{n} w_i}
    \label{eq:z-general-F} ~,
\end{equation}
where $n$ denotes the total number of rules, $i$ represents the $i$-th rule, $w_i$ is the antecedent activation (which corresponds to the MF output in our case), and $c_i$ is the consequent, representing the associated discomfort value. Equation~\ref{eq:z-general-F} encapsulates the complete formulation of discomfort as a function of height along the Z-axis. Properly defined MFs ensure a smooth and continuous transition across all height levels, maintaining consistency in discomfort estimation.



Since we use Gaussian membership functions, with a constant consequent and an antecedent output that directly corresponds to the membership function values, we can substitute $w$ with the Gaussian equation (Equation~\ref{eq:gaussian}). This results in an equivalent but more explicit formulation of the complete Z-axis discomfort function:
\begin{equation}
    f(z) = \frac{\sum_{k=1}^{n} e^{-\frac{(z - \mu_k)^2}{2\sigma_k^2}} c_k}{\sum_{k=1}^{n} e^{-\frac{(z - \mu_k)^2}{2\sigma_k^2}}}        ~,
    \label{eq:z-particular}
\end{equation}
%
%
%
where $k$ iterates over the six defined regions, with $c_k$ representing the discomfort value associated with each region.



\begin{figure}[htpb]
    \centering
    \includegraphics[width=0.50\linewidth]{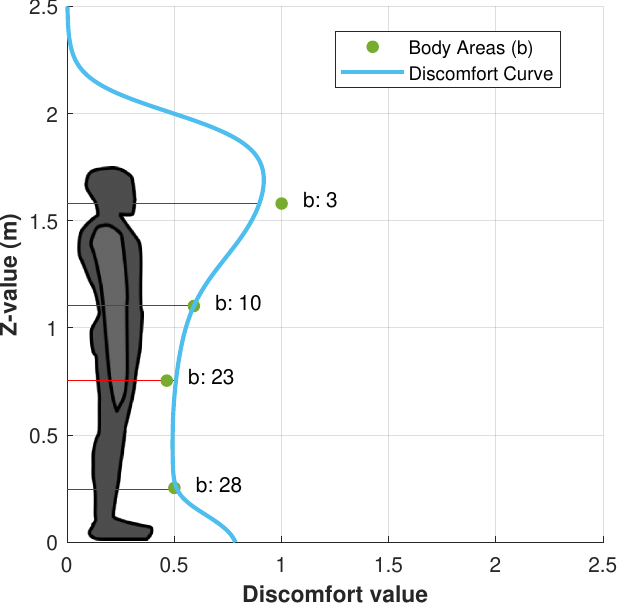}
    \caption{Z-axis discomfort function (blue curve), with green dots representing the most sensitive areas within each region, plotted by their corresponding $z$ and discomfort value.}
    \label{fig:person-graph}
    \vspace{-2mm}
\end{figure}

Figure~\ref{fig:person-graph} presents the discomfort values over the height range \([0, 2.5\,m]\) for a $1.75\,m$ tall person, with peak discomfort near the head and minimal discomfort around the legs.

When fine-tuning the Gaussian functions in our model, users should be aware that, in some configurations, discomfort values may increase again beyond a certain height due to the nature of fuzzy sets. To prevent unintended behavior, it may be necessary to manually set discomfort to zero above a specific threshold. If the goal is to extend the comfort zone above the head, replacing the top Gaussian membership function with an S-shaped function is recommended for a smoother decay.  
Additionally, while normalization is typically required when working exclusively with the Z-axis discomfort function, it is unnecessary at this stage since the function will later be unified with the XY discomfort function and normalized as a whole.

\subsection{Combining Discomfort in Z with XY}


After deriving the Z-axis discomfort function in Equation~\ref{eq:z-particular}, integrating it with an XY-plane discomfort model yields a comprehensive XYZ discomfort representation, defining the full 3D Personal Space.



We model the planar (XY) personal space using the traditional Asymmetric Gaussian Function (AGF)~\cite{Kirby2010}, widely applied in social robotics. Following Kirby’s approach, the AGF is centered at the individual's position $\mathbf{p} = (x, y)$ with orientation $\theta_i$ and is defined by variances $\sigma_h$, $\sigma_s$, and $\sigma_r$, representing the front, sides, and rear directions, respectively.

The parameters follow the values proposed in \cite{Kirby2010} for a stationary individual, as detailed below:
\begin{equation}
    \sigma_h = 1/2 \quad \textrm{and} \quad
    \sigma_s = \frac{2}{3} \sigma_h \quad \textrm{and} \quad
    \sigma_r = \frac{1}{2} \sigma_h ~.
    \nonumber
\end{equation}

\begin{figure}[htpb]
    \centering
    \includegraphics[width=0.65\linewidth]{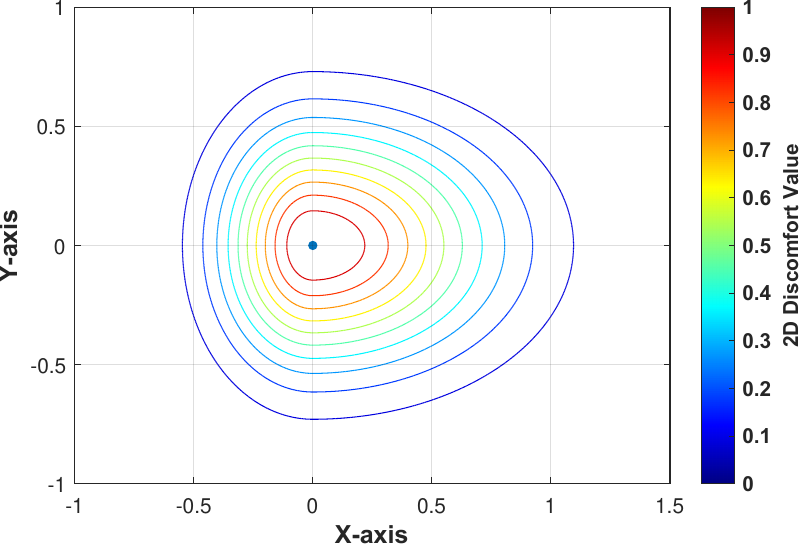}
    \caption{2D personal space for an individual at $\mathbf{q} = ((0, 0), 0)$, based on the model by Kirby~\cite{Kirby2010}, with $\sigma_h = 1/2$.}
    \label{fig:2d_contour}
\end{figure}

The AGF is elongated in the individual's facing direction by considering a linear velocity of $v = 1.0\,m/s$ in the model even for static individuals, highlighting increased discomfort levels along the body orientation, as shown in Figure~\ref{fig:2d_contour}.

Integrating the fuzzy Z-axis function with the AGF-based XY-plane model is straightforward, as both output discomfort values are within $[0,1]$. This combination follows an intuitive rationale: if an object is positioned at a height where the fuzzy Z-axis function returns a near-zero value, the overall discomfort remains negligible, regardless of its proximity in the XY-plane. Similarly, if an object is far in the XY-plane, leading to a low AGF output, the overall discomfort remains low regardless of height. In intermediate cases, the combined discomfort follows: when one component (Z or XY) is high and the other is intermediate, the resulting discomfort exceeds the intermediate level; if one is intermediate and the other low, the overall discomfort remains low.

To achieve this behavior, we use a geometric mean to combine both components. This approach naturally meets the described requirements, ensuring a smooth and consistent integration into a unified 3D proxemics discomfort function:
\begin{equation}
    \mathcal{S}(\mathcal{P}, \mathbf{r}) = \sqrt{\textrm{AGF}(\mathcal{P},\mathbf{r}_x, \mathbf{r}_y)*f(\mathbf{r}_z)}
    \label{eq:3D-complete} ~.
\end{equation}


Finally, a normalization step is applied to scale the maximum values of the AGF and $f(z)$ integration to 1. This guarantees consistency and preserves the relative contributions of both components in the final discomfort function.

\section{Results and Discussion}

\subsection{Illustrative example}



\begin{figure}[htpb]
    \centering
    \includegraphics[width=0.85\linewidth]{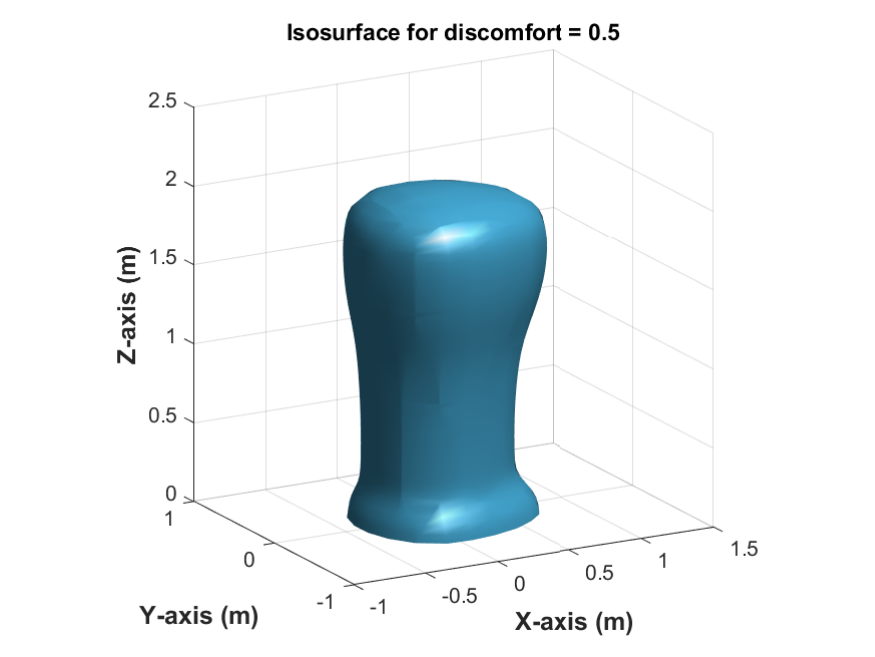}
    \caption{3D isosurface representing a discomfort threshold value of $0.5$ for a person of height 1.75 m, centered at (0,0) and facing along the X-axis. 
    }
    \label{fig:3D_result}
\end{figure}

\begin{figure}[htpb]
    \centering
    \includegraphics[width=\linewidth, trim= 100 300 100 250, clip]{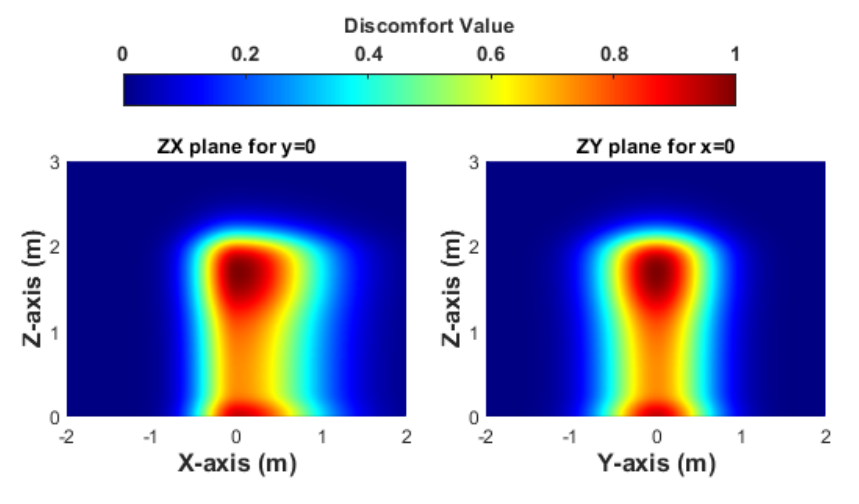}
    \caption{Surface plots of the XZ and YZ planes for a person at $\mathbf{q} = ((0, 0), 0)$ and with a height of $1.75\,m$.}
    \label{fig:surface_plot}
\end{figure}

To illustrate the spatial distribution of discomfort, we present the 3D isosurface corresponding to a discomfort threshold value of $0.5$ in Figure~\ref{fig:3D_result}. Additionally, Figure~\ref{fig:surface_plot} includes 2D surface plots of the ZX and ZY planes, where the color coding represents the corresponding discomfort values.


Discomfort values are highest near the head, reflecting greater sensitivity, and lowest around the legs, where sensitivity is typically lower. This reinforces the idea that robots should position components like arms or tools closer to the legs, rather than near the head, aligning with human comfort expectations. These findings emphasize the importance of considering both horizontal and vertical dimensions when designing robots to respect human personal space.




\subsection{Height variability}

\begin{figure}[htbp]
    \centering
    \subfigure[$h = 1.30\,m$]{
        \includegraphics[width=0.275\linewidth]{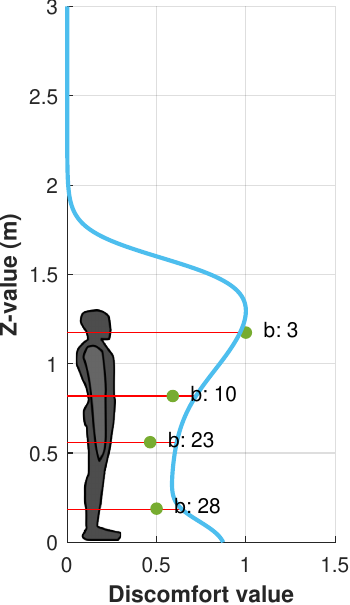}
        \label{fig:130cm}
    }
    \subfigure[$h = 1.75\,m$]{
        \includegraphics[width=0.275\linewidth]{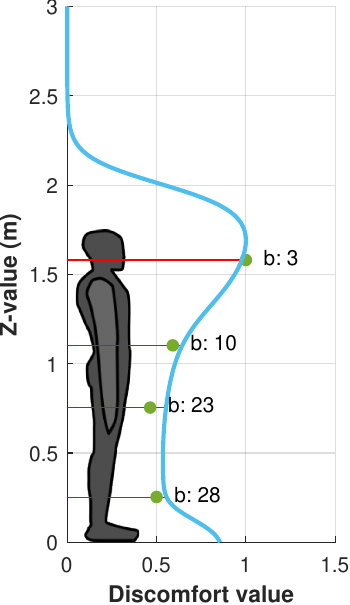}
        \label{fig:175cm}
    }    
    \subfigure[$h = 2.20\,m$]{
        \includegraphics[width=0.275\linewidth]{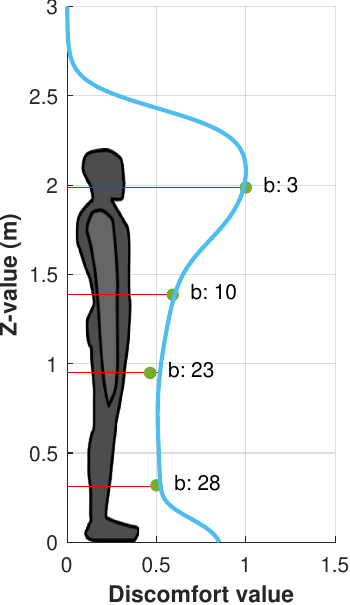}
        \label{fig:220cm}
    }    
    \caption{Adaptability to different height profiles.}
    \label{fig:height-comparison}
\end{figure}

Our model also adapts to varying heights, as illustrated in Figure~\ref{fig:height-comparison}, which presents normalized Z-axis discomfort profiles for individuals of $1.30\,m$, $1.75\,m$, and $2.20\,m$. The vertical distribution scales proportionally, ensuring proper alignment of key regions such as the head, chest, and lower body. This adaptability preserves consistency across diverse profiles without necessitating structural modifications.

Notably, the discomfort value approaches zero approximately $0.75\,m$ above the person's head. However, for flying robots, they would already prefer operating at a height around $h + 0.25\,m$, given our function, rather than below $h$. While the literature lacks consensus on this preference, recent studies suggest that flying above a person is favored for coexistence~\cite{Bretin2024}.

\section{Conclusion and Future Work}


This work introduced a novel 3D Proxemics model that extends traditional formulations by incorporating vertical discomfort, enabling a more comprehensive representation of human personal space. By integrating a Z-axis discomfort function with existing planar models, our approach provides a fundamental framework for respecting human comfort in all spatial dimensions. This advancement is key to enabling robots -- whether mobile, manipulators, or carrying objects -- to exhibit more socially aware behaviors in shared spaces.


In future work, we plan to validate our approach through real user studies to assess the alignment between predicted and actual discomfort levels, as well as to refine the model parameters. Additionally, we aim to integrate the model into real-time planning and control frameworks to enhance socially-aware robotic behavior in dynamic environments.





\bibliographystyle{IEEEtran}
\bibliography{bibliography}

\end{document}